\documentclass{article}

\usepackage{times}
\usepackage{graphicx} 
\usepackage{subfigure} 

\usepackage{natbib}

\usepackage{algorithm}
\usepackage{algorithmic}

\usepackage{hyperref}

\usepackage[accepted]{icml2016}
\usepackage[utf8]{inputenc} 
\usepackage{amsmath}
\usepackage{amssymb}
\usepackage{dsfont}
\usepackage{booktabs}       
\usepackage{tikz}
\usetikzlibrary{decorations.pathreplacing}
\usepackage{float}

\newtheorem{remark}{Remark}

\newcommand{\ie}{\emph{i.e.}{}}
\newcommand{\eg}{\emph{e.g.}{}}
\newcommand\iid{\ensuremath{\mathit{i.i.d.}}\ }

\newcommand{\wrt}{\emph{w.r.t.}{}}
\newcommand{\crit}{\mathcal{C}}

\newcommand{\st}{\emph{s.t.}{}}
\newcommand{\cdf}{\emph{c.d.f.}{}}

\def\mb{\mathbf}

\def\rset{\mathbb{R}}
\def\leb{\text{Leb}}
\def\argmin{\operatornamewithlimits{arg\,min}}

\def\argmin{\operatornamewithlimits{arg\,min}}

\icmltitlerunning{Evaluation of Unsupervised Anomaly Detection Algorithms}

\begin{document} 

\twocolumn[
\icmltitle{How to Evaluate the Quality of Unsupervised Anomaly Detection Algorithms?}


\icmlauthor{Nicolas Goix}{nicolas.goix@telecom-paristech.fr}
\icmladdress{LTCI, CNRS, Télécom ParisTech, Université Paris-Saclay, 75013, Paris, France}

\icmlkeywords{One-Class Classification, Anomaly Detection, evaluation criterion}

\vskip 0.3in
]

\begin{abstract} 
When sufficient labeled data are available, classical criteria based on \emph{Receiver Operating Characteristic} (ROC) or \emph{Precision-Recall} (PR) curves can be used to compare the performance of unsupervised anomaly detection algorithms. However, in many situations, few or no data are labeled. This calls for alternative criteria one can compute on non-labeled data. In this paper, two criteria that do not require labels are empirically shown to discriminate accurately (\wrt~ROC or PR based criteria) between algorithms. 
These criteria are based on existing Excess-Mass (EM) and Mass-Volume (MV) curves, which generally cannot be well estimated in large dimension.
A methodology based on feature sub-sampling and aggregating is also described and tested, extending the use of these criteria to high-dimensional datasets and solving major drawbacks inherent to standard EM and MV curves.
\end{abstract} 

\section{Introduction}
\label{sec:intro}
When labels are available, classical ways to evaluate the quality of an anomaly scoring function are the ROC and PR curves.
Unfortunately, most of the time, data come without any label. In lots of industrial setups, labeling datasets calls for costly human expertise, while more and more unlabeled data are available.
A huge practical challenge is therefore to have access to criteria able to discriminate between unsupervised algorithms without using any labels. 
In this paper, we formalize and justify the use of two such criteria designed for unsupervised anomaly detection (AD), and adapt them to large dimensional data. Strong empirical performance demonstrates the relevance of our approach. 


The common underlying assumption behind AD is that anomalies occur in low probability regions of the data generating process.
This formulation motivates many statistical AD methods. 
Classical parametric techniques \cite{Barnett94, Eskin2000} assume that the normal data are generated by a distribution belonging to some  specific and \emph{a priori} known parametric model.  
The most popular non-parametric approaches include algorithms based on density (level set) estimation \cite{Scholkopf2001, Scott2006, Breunig2000LOF}, on dimensionality reduction \cite{Shyu2003, Aggarwal2001} or on decision trees \cite{Liu2008}.
One may refer to \cite{Hodge2004survey, Chandola2009survey, Patcha2007survey, Markou2003survey} for overviews of current research on AD.
It turns out that the overwhelming majority of AD algorithms return more than a binary label, normal/abnormal. They first compute a \emph{scoring function}, which is converted to a binary prediction, typically by imposing some threshold based on its statistical distribution.


\textbf{What is a scoring function?}
As anomalies are very rare, their structure cannot be observed in the data, in particular their distribution. 
It is common and convenient to assume that anomalies occur in the tail of $F$ the distribution of normal data, so that the goal is to estimate density level sets of $F$.
%
This setup is typically the one of the One-Class Support Vector Machine (OneClassSVM) algorithm developped in \cite{Scholkopf2001}, which extends the SVM methodology \cite{Shawe2004} to handle training using only positive information.
The underlying assumption is that we observe data in $\rset^d$ from the normal class only, with underlying distribution $F$ and underlying density $f: \rset^d \to \rset$. The goal is to estimate density level sets $(\{\mb x, f( \mb x) > t\})_{t>0}$ with $t$ close to $0$.
Such estimates are encompassed into a \emph{scoring function}: any measurable function $s:~\rset^d \to \rset_+$ integrable \wrt~the Lebesgue measure $\leb(.)$, whose level sets are estimates of the level sets of the density. 
Any scoring function defines a preorder on $\rset^d$ and thus a ranking on a set of new observations. This ranking can be interpreted as a degree of abnormality, the lower $s(x)$, the more abnormal $x$. 

\textbf{How to know if a scoring function is good?}
How can we know if the preorder induced by a scoring function $s$ is `close' to that of $f$, or equivalently if these induced level sets are close to those of $f$? 
The problem is to define this notion of proximity into a criterion $\crit$, optimal scoring functions $s^*$ being then defined as those optimizing $\crit$. 
It turns out that for any strictly increasing transform $T: \text{Im(f)} \to \mathbb{R} $, the level sets of $T\circ f$ are exactly those of $f$. Here and hereafter, $\text{Im(f)} $ denotes the image of the mapping $f$. For instance, $2f$ or $f^2$ are perfect scoring functions, just as $f$. Thus, we cannot simply consider a criterion based on the distance of $s$ to the true density, \eg~$\crit(s)=\|s - f\|$.
We seek for a similar criterion which is invariant by increasing transformation of the output $s$. In other words, the criterion should be defined in such a way that the collection of level sets of an optimal scoring function $s^*(x)$ coincides with that related to $f$. Moreover, any increasing transform of the density should be optimal regarding $\crit$.

In the litterature, two functional criteria admissible \wrt~these requirements have been introduced: the Mass-Volume (MV) \cite{CLEM13} and the Excess-Mass (EM) \cite{AISTAT15} curves.
Formally, it allows to consider $\crit^{\Phi}(s) = \| \Phi(s) - \Phi(f) \|$ (instead of $\|s - f\|$) 
with $\Phi: \mathbb{R} \to \mathbb{R}_+$ verifying $\Phi(T \circ s) = \Phi(s)$ 
for any scoring function $s$ and increasing transform $T$. Here $\Phi(s)$ denotes either the mass-volume curve $MV_s$ of $s$ or its excess-mass curve $EM_s$, which are defined in the next section.  
While such quantities have originally been introduced to build scoring functions \emph{via}
Empirical Risk Minimization (ERM), the MV-curve has been used recently for the calibration of the One-Class SVM \cite{Thomas2015}.
When used to attest the quality of some scoring function, the volumes induced become unknown and must be estimated, which is challenging in large dimension.

In this paper, we define two numerical performance criteria based on MV and EM curves, which are tested \wrt three classical AD algorithms. A wide range on real labeled datasets are used in the benchmark. In addition, we propose a method based on feature sub-sampling and aggregating. It allows to scale 
 this methodology to high-dimensional data which we use on the higher-dimensional datasets. We compare the results to ROC and PR criteria, which use the data labels hidden to MV and EM curves. 

 This paper is structured as follows. Section~\ref{background} introduces EM and MV curves and defines associated numerical criteria. In Section~\ref{scaling-dim}, the feature sub-sampling based methodology to extend their use to high dimension is described. Finally, experiments on a wide range of real datasets are provided in Section~\ref{sec:benchmarks}.

\section{Mass-Volume and Excess-Mass based criteria}
\label{background}
We place ourselves in a probability space $(\Omega, \mathcal{F}, \mathbb{P})$. We observe $n$ \iid~realizations $\mb X_1,\ldots,\mb X_n$ of a random variable $\mb X:\Omega \to \mathbb{R}^d$ representing the normal behavior, with \cdf~$F$ and density $f$ \wrt~the Lebesgue measure on $\mathbb{R}^d$. We denote by $\mathcal{S}$ the set of all scoring functions, namely any measurable function $s:~\rset^d \to \rset_+$ integrable \wrt~the Lebesgue measure.
We work under the assumptions that the density $f$ 
has no flat parts and is bounded. Excess-Mass and Mass-Volume curves are here introduced in a different way they originally were in \cite{CLEM13, AISTAT15}. We use equivalent definitions for them since 
the original definitions were more adapted to the ERM paradigm than to the issues adressed here.

\textbf{Preliminaries.}
Let $s\in \mathcal{S}$ be a scoring function. In this context \cite{CLEM13,AISTAT15}, the MV and EM curves of $s$ can be written as
\noindent
\begin{align}
\label{MV-def}
 MV_s(\alpha) &= \inf_{u \ge 0}~~ \leb(s \ge u) ~~\st~~\mathbb{P}(s(\mb X) \ge u) \ge \alpha\\
\label{EM-def}
 EM_s(t) &= \sup_{u \ge 0}~~ \mathbb{P}(s(\mb X) \ge u) ~-~ t \leb(s \ge u)
\end{align}
for any $\alpha\in (0,1)$ and $t >0$.
The optimal curves are $MV^* = MV_f = MV_{T \circ f}$ and $EM^* = EM_f = EM_{T \circ f}$ for any increasing transform $T: \text{Im(f)} \to \mathbb{R}$. 
It can be proven \cite{CLEM13, AISTAT15} that for any scoring function $s$, $MV^*(\alpha)\leq MV_s(\alpha)$ for all $\alpha\in (0,1)$ and $EM^*(t) \ge EM_s(t)$ for all $t > 0$.

\textbf{Numerical unsupervised criteria.}
The main advantage of EM compared to MV is that the area under its curve (AUC) is finite, even if the support of the distribution $F$ is not.
As curves cannot be trivially compared, consider the $L^1$-norm $\|.\|_{L^1(I)}$ with $I\subset \rset$ an interval. As $MV^*=MV_f$ is below $MV_s$ pointwise, $\argmin_s \| MV_s - MV^* \|_{L^1(I)} = \argmin \| MV_s \|_{L^1(I)} $. We thus define
$\crit^{MV}(s) = \| MV_s \|_{L^1(I^{MV})},$ which is equivalent to consider $\| MV_s - MV^* \|_{L^1(I^{MV})}$ as mentioned in the introduction. As we are interested in evaluating accuracy on large density level-sets, one natural interval $I^{MV}$ would be for instance $[0.9, 1]$. However, MV diverges in $1$ when the support is infinite, so that we arbitrarily take $I^{MV} = [0.9, 0.999].$
The smaller is $\crit^{MV}(s)$, the better is the scoring function $s$.
Similarly, we consider $\crit^{EM}(s) = \| EM_s \|_{L^1(I^{EM})}, $ this time considering $I^{EM} = [0,EM^{-1}(0.9)],$ with $EM_s^{-1}(0.9) := \inf\{t\ge 0,~ EM_s(t) \le 0.9\}$, as $EM_s(0)$ is finite (equal to $1$). We point out that such small values of $t$ correspond to large level-sets. Also, we have observed that
$EM_s^{-1}(0.9)$ (as well as $EM_f^{-1}(0.9)$) varies significantly depending on the dataset. Generally, for datasets in large dimension, it can be 
very small (in the experiments, smallest values are of order $10^{-7}$) as it is of the same order of magnitude as the inverse of the total support volume.

\textbf{Estimation.} 
As the distribution $F$ of the normal data is generally unknown, MV and EM curves must be estimated. Let $s\in \mathcal{S}$ and $\mb X_1,\; \ldots,\; \mb X_n$ be an i.i.d. sample with common distribution $F$ and set $\mathbb{P}_n(s \ge t)=\frac{1}{n}\sum_{i=1}^n\mathds{1}_{s(\mb X_i)\geq t}.$ The empirical MV and EM curves of $s$ are then simply defined as empirical version of \eqref{MV-def} and \eqref{EM-def}, 
\begin{align}
\label{MV-def-emp}
\widehat{MV}_s(\alpha) = \inf_{u \ge 0} \left\{ \leb(s \ge u) ~~\st~ \mathbb{P}_n(s \ge u) \ge \alpha \right\}
\end{align}
\begin{align}
\label{EM-def-emp}
\widehat{EM}_s(t) = \sup_{u \ge 0} \mathbb{P}_n(s \ge u) ~-~ t \leb(s \ge u)
\end{align}
Note that in practice, the volume $\leb(s \ge u)$ is estimated using Monte-Carlo approximation, which only applies to small dimensions.
Finally, we obtain the empirical EM and MV based performance criteria:
\begin{align}
\label{eq:standard_emp_EM}\widehat{\crit}^{EM}(s) &= \| \widehat{EM}_s \|_{L^1(I^{EM})}  &&I^{EM} = [0,\widehat{EM}^{-1}(0.9)],\\
\label{eq:standard_emp_MV}\widehat{\crit}^{MV}(s) &= \| \widehat{MV}_s \|_{L^1(I^{MV})}  &&I^{MV} = [0.9, 0.999].
\end{align}
\section{Scaling with dimension}
\label{scaling-dim}
In this section we propose a methodology to scale the use of the EM and MV criteria to large dimensional data. It consists in sub-sampling training \emph{and} testing data along features, thanks to a parameter $d'$ controlling the number of features randomly chosen for computing the (EM or MV) score. Replacement is done after each draw of features $F_1,\ldots,F_{m}$. A partial score $\widehat \crit_k^{MV}$ (resp. $\widehat \crit_k^{EM}$) is computed for each draw $F_k$ using \eqref{eq:standard_emp_EM} (resp. \eqref{eq:standard_emp_MV}). The final performance criteria are obtained by averaging these partial criteria along the different draws of features. This methodology is described in Algorithm~\ref{algo:EMMV}.
\begin{algorithm}[!tbh]
\caption{~~Evaluate AD algo. on high-dimensional data}
\label{algo:EMMV}
\begin{algorithmic}
  \STATE \textbf{Inputs}: AD algorithm $\mathcal{A}$, data set $X = (x^j_i)_{1 \le i \le n, 1 \le j \le d }$, feature sub-sampling size $d'$, number of draws $m$.
  \FOR{$k=1,\ldots,m$}
    \STATE{randomly select a sub-group $F_k$ of $d'$ features}
    \STATE{compute the associated scoring function $\widehat s_{k} = \mathcal{A}\big((x^j_i)_{1 \le i \le n,~j \in F_k}\big)$}
    \STATE compute $\widehat{\crit}_k^{EM} = \| \widehat{EM}_{\widehat s_k} \|_{L^1(I^{EM})}$ using \eqref{eq:standard_emp_EM} or $\widehat{\crit}_k^{MV} = \| \widehat{MV}_{\widehat s_k} \|_{L^1(I^{MV})}$ using \eqref{eq:standard_emp_MV}
  \ENDFOR 
  \STATE \textbf{Return} performance criteria: $$\widehat{\crit}^{EM}_{high\_dim} (\mathcal{A})= \frac{1}{m} \sum_{k=1}^m\widehat \crit_k^{EM} \text{~~~~(idem for MV)}$$
\end{algorithmic}
\end{algorithm}
A drawback from this approach is that we do not evaluate combinations of more than $d'$ features within the dependence structure. However, according to our experiments, this is enough in most of the cases. Besides, we solve two major drawbacks inherent to MV or EM criteria, 
which come from the Lebesgue reference measure:
\textbf{1)}  EM or MV performance criteria cannot be estimated in large dimension,
\textbf{2)}  EM or MV performance criteria cannot be compared when produced from spaces of different dimensions. 
\begin{remark}({\sc Feature Importances})
With standard MV and EM curves, the benefit of using or not some feature $j$ in training \emph{cannot} be evaluated, since reference measures of $\mathbb{R}^d$ and $\mathbb{R}^{d+1}$ cannot be compared.
Solving the second drawback precisely allows to evaluate the importance of features.
By sub-sampling features, we can compare accuracies with or without using feature $j$: when computing $\widehat{\crit}^{MV}_{high\_dim}$ or $\widehat{\crit}^{EM}_{high\_dim}$ using Algorithm~\ref{algo:EMMV}, this is reflected in the fact that $j$ can (resp. cannot) be drawn.
\end{remark}
Remarks on theoretical grounds and default parameters are provided in supplementary material.
\section{Benchmarks}
\label{sec:benchmarks}
\textbf{Does performance in term of EM/MV correspond to performance in term of ROC/PR?} Can we recover, on some fixed dataset and without using any labels, which algorithm is better than the others (according to ROC/PR criteria)?
In this section we study four different empirical evaluations (ROC, PR, EM, MV) of three classical state-of-the-art AD algorithms, One-Class SVM  \cite{Scholkopf2001}, Isolation Forest \cite{Liu2008}, and Local Outlier Factor (LOF) algorithm \cite{Breunig2000LOF}, on 12 well-known AD datasets. Two criteria use labels (ROC and PR based criteria) and two do not (EM and MV based criteria).
For ROC and PR curves, we consider the area under the (full) curve (AUC). For the excess-mass curve $EM(t)$ (resp. mass-volume curve), we consider the area under the curve on the interval $[0, EM^{-1}(0.9)]$ (resp. $[0.9, 0.999]$) as described in Section~\ref{background}.
A full description of the datasets is available in supplementary material.
The experiments are performed both in a novelty detection framework (also named semi-supervised framework, the training set consisting of normal data only) and in an unsupervised framework (the training set is polluted by anormal data).
In the former case, we simply removed anomalies from the training data, and EM and PR criteria are estimated using only normal data.
In the latter case, the anomaly rate is arbitrarily bounded to $10\%$ max, and EM and PR criteria are estimated with the same test data used for ROC and PR curves, without using their labels.
\begin{table*}[!ht]
\centering
\caption{Results for the novelty detection setting. One can see that ROC, PR, EM, MV often do agree on which algorithm is the best (in bold), which algorithm is the worse (underlined) on some fixed datasets. When they do not agree, it is often because ROC and PR themselves do not, meaning that the ranking is not clear.}
\label{table:results-semisupervised}
\footnotesize
\tabcolsep=0.11cm
\resizebox{0.9 \linewidth}{!} {
\begin{tabular}{l cccc c cccc c cccc}
\toprule
Dataset      & \multicolumn{4}{c}{iForest}& & \multicolumn{4}{c}{OCSVM}&  & \multicolumn{4}{c}{LOF} \\ 
  \cmidrule{1-15}

~            & ROC  & PR   & EM    &  MV  &  & ROC  & PR   & EM    & MV     &  & ROC  & PR   & EM    & MV    \\
adult        &\bf 0.661 &\bf 0.277 &\bf 1.0e-04&\bf 7.5e01&  &0.642 &0.206 &2.9e-05& 4.3e02 &  &\underline{0.618} &\underline{0.187}&\underline{1.7e-05}&\underline{9.0e02} \\
http         &0.994 &0.192 &1.3e-03&9.0   &  &\bf 0.999 &\bf 0.970 &\bf 6.0e-03&\bf 2.6  &     &\underline{0.946} &\underline{0.035} &\underline{8.0e-05}&\underline{3.9e02} \\
pima         &0.727 &0.182 &5.0e-07&\bf 1.2e04&  &\bf 0.760 &\bf 0.229 &\bf 5.2e-07&\underline{1.3e04} &   &\underline{0.705} &\underline{0.155} &\underline{3.2e-07}&2.1e04 \\
smtp         &0.907 &\underline{0.005} &\underline{1.8e-04}&\underline{9.4e01}&  &\underline{0.852} &\bf 0.522 &\bf 1.2e-03&8.2    &   &\bf 0.922 &0.189 & 1.1e-03&\bf 5.8    \\
wilt         &0.491 &0.045 &4.7e-05&\underline{2.1e03} & &\underline{0.325} &\underline{0.037} &\bf 5.9e-05&\bf 4.5e02 &   &\bf 0.698 &\bf 0.088 &\underline{2.1e-05}&1.6e03 \\ 
 &&&&&&&&&&&&&& \\
annthyroid   &\bf 0.913 &\bf 0.456 &\bf 2.0e-04&2.6e02 & &\underline{0.699} &\underline{0.237} &\underline{6.3e-05}&\bf 2.2e02 &   &0.823 &0.432 &6.3e-05&\underline{1.5e03} \\
arrhythmia   &\bf 0.763 &\bf 0.487 &\bf 1.6e-04&\bf 9.4e01 & &0.736 &0.449 &1.1e-04&1.0e02   & &\underline{0.730} &\underline{0.413} &\underline{8.3e-05}&\underline{1.6e02} \\
forestcov.   &\underline{0.863} &\underline{0.046} &\underline{3.9e-05}&\underline{2.0e02}&  &0.958 &0.110 &5.2e-05&1.2e02  &  &\bf 0.990 &\bf 0.792 &\bf 3.5e-04&\bf 3.9e01 \\
ionosphere   &\underline{0.902} &\underline{0.529} &\underline{9.6e-05}&\underline{7.5e01} & &\bf 0.977 &\bf 0.898 &\bf 1.3e-04&\bf 5.4e01  &  &0.971 &0.895 &1.0e-04&7.0e01 \\
pendigits    &0.811 &0.197 &2.8e-04&2.6e01 & &\underline{0.606} &\underline{0.112} &\underline{2.7e-04}&\underline{2.7e01}   & &\bf 0.983 &\bf 0.829 &\bf 4.6e-04&\bf 1.7e01 \\
shuttle      &0.996 &0.973 &1.8e-05&5.7e03 & &\underline{0.992} &\underline{0.924} &\bf 3.2e-05&\bf 2.0e01   & &\bf 0.999 &\bf 0.994 &\underline{7.9e-06}&\underline{2.0e06} \\
spambase     &\bf 0.824 &\bf 0.371 &\bf 9.5e-04&\bf 4.5e01&  &\underline{0.729} &0.230 &4.9e-04&1.1e03  &  &0.754 &\underline{0.173} &\underline{2.2e-04}&\underline{4.1e04} \\
\bottomrule
\end{tabular}
}
\end{table*}
Recall that standards EM and MV performance criteria refering on the Lebesgue measure, they require volume estimation. They only apply to continuous datasets, with small dimension ($d \le 8$). The datasets verifying these requirements are \emph{http}, \emph{smtp}, 
\emph{pima}, \emph{wilt} and \emph{adult}.
For the other datasets, we use the performance criteria $\widehat{\crit}^{MV}_{high\_dim}$ and $\widehat{\crit}^{EM}_{high\_dim}$ computed with Algorithm~\ref{algo:EMMV}. We arbitrarily chose $m = 50$ and $d'=5$, which means that $50$ draws of $5$ features, with replacement after each draw, are done.
Other parameters have also been tested but are not presented here. This default parameters are a compromise between computational time and performance, in particular on the largest dimensional datasets. The latter require a relatively large product $m \times d'$, which is the maximal number of different features that can be drawn.

EM, MV, ROC and PR curves AUCs are presented in Table~\ref{table:results-semisupervised} for the novelty detection framework. Additional figures and results for the unsupervised framework are available in supplementary material. 
%
%
Results from Table~\ref{table:results-semisupervised} can be summarized as follows. Consider the $36$ possible pairwise comparisons between the three algorithms over the twelve datasets
\begin{align}
\label{eq:pairs}
\nonumber  \big\{\big(A_1 \text{~on~} \mathcal{D}, A_2 \text{~on~} \mathcal{D}\big),&~ A_{12} \in \{\text{iForest, LOF, OCSVM}\}, \\ & \mathcal{D} \in \{\text{adult, \ldots, spambase}\} \big\}.
\end{align}
For each dataset $\mathcal{D}$, there are three possible pairs (iForest on $\mathcal{D}$, LOF on $\mathcal{D}$), (OCSVM on $\mathcal{D}$, LOF on $\mathcal{D}$) and (OCSVM on $\mathcal{D}$, iForest on $\mathcal{D}$).
Then the EM-score discriminates $28$ of them ($78\%$) as ROC score does, and $29$ ($81\%$) of them as PR score does. 
Intuitively this can be interprated as follows. Choose randomly a dataset $\mathcal{D}$ among the twelve available, and two algorithms $A_1$, $A_2$ among the three available. This amounts to choose at random a pairwise comparison ($A_1$ on $\mathcal{D}$, $A_2$ on $\mathcal{D}$) among the $36$ available. Suppose that according to ROC criterion, $A_1$ is better than $A_2$ on dataset $\mathcal{D}$, \ie~($A_1$ on $\mathcal{D}$) $\succ$ ($A_2$ on $\mathcal{D}$). Then the EM-score discriminates $A_1$ and $A_2$ on dataset $\mathcal{D}$ in the same way, \ie~also finds $A_1$ to be better than $A_2$ on dataset $\mathcal{D}$, this with $78$ percent chance.

 Besides, let us consider pairs ($A_1$ on $\mathcal{D}$, $A_2$ on $\mathcal{D}$) which are
similarly ordered by ROC and PR criteria, namely \st~ $A_1$ is better than $A_2$ (or the reverse) on dataset $\mathcal{D}$ according to both EM and PR. According to Table~\ref{table:results-semisupervised}, this represents every pairs but one in \emph{spambase} and two in \emph{smtp}. Then, one achieves $27/33 = 82\%$ of similarly discriminated pairs (\wrt~to ROC and PR criteria). Moreover, EM is able to recover the exact (\wrt~ROC and PR criteria) ranking of ($A_1$ on $\mathcal{D}$, $A_2$ on $\mathcal{D}$, $A_3$ on $\mathcal{D}$) on every datasets $\mathcal{D}$ excepting \emph{wilt} and \emph{shuttle}. For \emph{shuttle}, note that ROC scores are very close to each other ($0.996$, $0.992$, $0.999$) and thus not clearly discriminates algorithms. The only significant error committed by EM is for the \emph{wilt} dataset (on which no feature sub-sampling is done due to the low dimension). This may come from anomalies not being far enough in the tail of the normal distribution, \eg~forming a cluster near the support of the latter distribution. 

Same conclusions and similar accuracies hold for MV-score, which only makes one additional error on the pair (iForest on $pima$, OCSVM on $pima$).
Considering all the 36 pairs \eqref{eq:pairs}, 
one observes $75\%$ of good comparisons \wrt~ROC-score, and $72\%$ \wrt~PR score.
Considering the pairs which are similarly ordered by ROC and PR criteria, this rate increases to $25/33 = 76\%$. The errors are essentially made on \emph{shuttle}, \emph{wild} and \emph{annthyroid} datasets. 

To conclude, when one algorithm has 
better performance than another on some fixed dataset, according to both ROC and PR AUCs, one can expect to recover it without using labels with an accuracy of $82\%$ in the novelty detection framework (and $77\%$ in the unsupervised framework, cf. supplementary material).
%
%

\section{Conclusion}
We (almost) do not need labels to evaluate anomaly detection algorithms (on continuous data). According to our benchmarks, the EM and MV based numerical criteria introduced in this paper are (in approximately $80$ percent of the cases) able to recover which algorithm is better than the other on some dataset (with potentially large dimensionality), without using labels. High-dimensional datasets are dealt with using a method based on feature sub-sampling. This method also brings flexibility to EM and MV criteria, allowing for instance to evaluate the importance of features.

\bibliographystyle{icml2016}
\bibliography{Thesis}

\clearpage
\section*{Supplementary Material}

\subsection{additional intuition behind EM/MV}
Note that $MV^*(\alpha)$ is the optimal value of the constrained minimization problem
\begin{equation}\label{eq:MV}\min_{\Gamma~ \text{borelian}} ~\leb(\Gamma) ~~~\st~~ \mathbb{P}(\mb X \in \Gamma) \ge \alpha.
\end{equation}
The minimization problem \eqref{eq:MV} has a unique solution $\Gamma_\alpha^*$ of mass $\alpha$ exactly, referred to as \textit{minimum volume set} \cite{Polonik97}: $MV^*(\alpha)=\leb(\Gamma^*_\alpha)$ and $\mathbb{P}(\mb X \in \Gamma_\alpha^*)=\alpha$. 

Similarly, the optimal EM curve is linked with the notion of density excess-mass (as introduced in the seminal contribution \cite{Polonik95}). The main idea is to consider a Lagrangian formulation of the constrained minimization problem obtained by exchanging constraint and objective in \eqref{eq:MV},
\begin{align}
\label{eq:EM}
EM^*(t):=\max_{\Omega\text{ borelian} } \{ {\mathbb{P}} (\mb X\in \Omega)-t\leb(\Omega) \}.
\end{align}
Figure~\ref{aistat:MVcurve} compares the mass-volume and excess-mass approaches.
\begin{center}
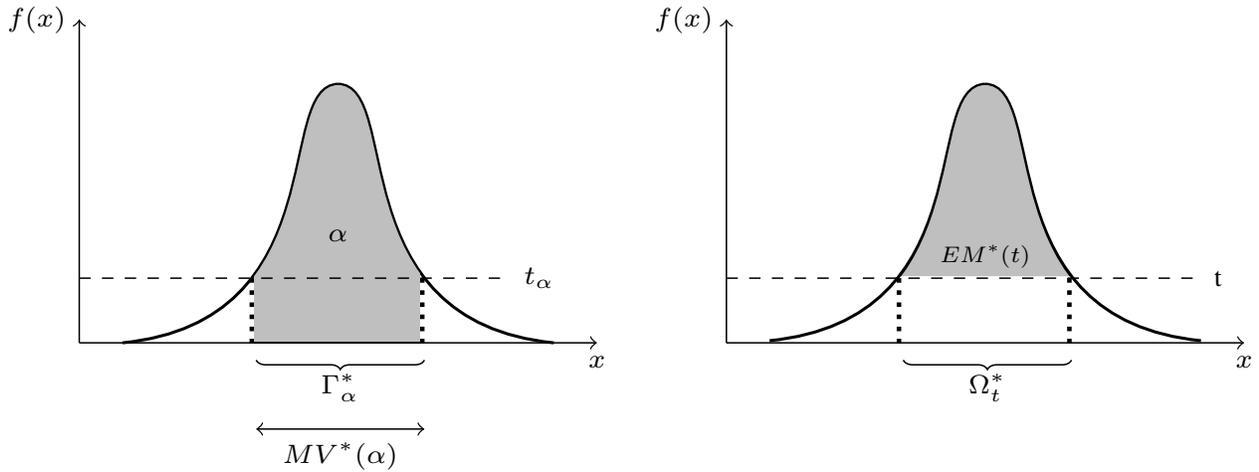
\begin{figure*}[!ht]
\caption{Comparison between $MV^*(\alpha)$ and $EM^*(t)$}
\label{aistat:MVcurve}
\centering
\resizebox{\linewidth}{!} {
\begin{tikzpicture}[scale=2.]

\draw[->](7.8,0)--(10.2,0) node[below]{\scriptsize $x$};
\draw[->](7.8,0)--(7.8,1.5) node[left]{\scriptsize $f(x)$}; 
\draw [thick] (8,0) ..controls +(1,0.1) and +(-0.3,-0.01).. (9,1.2);
\draw [thick] (9,1.2) ..controls +(0.3,-0.01) and +(-1,0.1).. (10,0);

\draw[dotted,very thick](8.60,0.3)--(8.60,0) node[right]{};
\draw[dotted,very thick](9.39,0.3)--(9.39,0) node[right]{};


\begin{scope} 
\clip (8.61,0)--(9.38,0)--(9.38,1.5)--(8.61,1.5)--(8.61,0) ; 

\path[draw,fill=lightgray] (8,0) ..controls +(1,0.1) and +(-0.3,-0.01).. (9,1.2)--
(9,1.2) ..controls +(0.3,-0.01) and +(-1,0.1).. (10,0)--
(8,0)--(10,0) --cycle;
\end{scope}

\draw[dashed](7.8,0.3)--(9.8,0.3) node[right]{\scriptsize $t_\alpha$};

\draw[decorate,decoration={brace}]
(9.4,-0.08)--(8.62,-0.08) node[below,pos=0.5] {\scriptsize $\Gamma_\alpha^*$};

\draw[<->]
(9.4,-0.4)--(8.62,-0.4) node[below,pos=0.5] {\scriptsize $MV^*(\alpha)$};

\draw (9,0.5) node[thick]{\scriptsize $\alpha$} ;

\draw[->](10.8,0)--(13.2,0) node[below]{\scriptsize $x$};
\draw[->](10.8,0)--(10.8,1.5) node[left]{\scriptsize $f(x)$}; 
\draw [thick] (11,0.01) ..controls +(1,0.1) and +(-0.3,-0.01).. (12,1.2);
\draw [thick] (12,1.2) ..controls +(0.3,-0.01) and +(-1,0.1).. (13,0.01);
\draw[dashed](10.8,0.3)--(13,0.3) node[right]{\scriptsize t};
\draw[dotted,very thick](11.60,0.3)--(11.60,0) node[right]{};
\draw[dotted,very thick](12.39,0.3)--(12.39,0) node[right]{};


\begin{scope} 
\clip (11,0.31)--(13,0.308)--(13,1.5)--(11,1.5)--(11,0.308) ; 
\path[draw,fill=lightgray] (11,0) ..controls +(1,0.1) and +(-0.3,-0.01).. (12,1.2)--
(12,1.2) ..controls +(0.3,-0.01) and +(-1,0.1).. (13,0)--
(11,0.3)--(13,0.3) --cycle;
\end{scope}

\draw[decorate,decoration={brace}]
(12.4,-0.08)--(11.62,-0.08) node[below,pos=0.5] {\scriptsize  $\Omega_t^*$};

\draw (12,0.4) node[scale=0.85]{\scriptsize $EM^*(t)$} ;
\draw (11.8,-0.5) node[thick]{ } ;
\end{tikzpicture}
}
\end{figure*}
\end{center}

\begin{remark}({\sc Link with ROC curve})
To evaluate unsupervised algorithms, it is common to generate uniform outliers and then use the ROC curve approach. Up to identify the Lebesgue measure of a set to its empirical version (\ie~the proportion of uniform point inside),
this approach is equivalent to using the mass-volume curve \cite{CLEM14}.
 However, in the former approach, the volume estimation does not appear directly, so that the (potentially huge) amount of uniform points needed to provide a good estimate of a volume is often not respected, yielding optimistic performances.
\end{remark}

\subsection{Remarks on the feature sub-sampling based Algorithm 1.}

\begin{remark}({\sc Theoretical Grounds})
Criteria $\widehat{\crit}^{MV}_{high\_dim}$ or $\widehat{\crit}^{EM}_{high\_dim}$ do not evaluate a specific scoring function $s$ produced by some algorithm (on some dataset), but the algorithm itself \wrt~the dataset at stake. Indeed, these criteria proceed with the average of partial scoring functions on sub-space of $\mathbb{R}^d$. We have no theoretical guaranties that the final score does correspond to some scoring function defined on $\mathbb{R}^d$. In this paper, we only show that from a practical point of view, it is a useful and accurate methodology to compare algorithms performance on large dimensional datasets.
\end{remark}
\begin{remark}({\sc Default Parameters}) 
In our experiments, we arbitrarily chose $m = 50$ and $d'=5$. This means that $50$ draws of $5$ features (with replacement after each draw) have been done. Volume in spaces of dimension $5$ have thus to be estimated (which is feasible with Monte-Carlo), and $50$ scoring functions (on random subspaces of dimension $5$) have to be computed by the algorithm we want to evaluate.
The next section shows (empirically) that these parameters achieve a good accuracy on the collection of datasets studied, the largest dimension considered being $164$.
\end{remark}

\subsection{Datasets description}

\begin{table*}[!ht]
\caption{Original Datasets characteristics}
\label{table:data}
\centering
\resizebox{0.9 \linewidth}{!} {
\begin{tabular}{l cc ll }
  \toprule
  ~           & nb of samples      & nb of features     & ~~~~~~~~~~~~~~~~~~~~~~~~~anomaly class      & ~                  \\ \cmidrule{1-5}
  adult       & 48842              & 6                  &    class '$>50K$'                           &      (23.9\%)      \\
  http        & 567498             & 3                  &      attack                                 &    (0.39\%)        \\
  pima        & 768                & 8                  &    pos (class 1)                            &        (34.9\%)    \\
  smtp        & 95156              & 3                  &      attack                                 &    (0.03\%)        \\
  wilt        & 4839               & 5                  &    class 'w' (diseased trees)               &    (5.39\%)        \\
  annthyroid  & 7200               & 6                  &    classes $\neq$ 3                         &        (7.42\%)    \\
  arrhythmia  & 452                & 164                &    classes $\neq$ 1 (features 10-14 removed)&  (45.8\%)          \\
  forestcover & 286048             & 10                 &    class 4  (vs. class 2 )                  &           (0.96\%) \\
  ionosphere  & 351                & 32                 &    bad                                      &       (35.9\%)     \\
  pendigits   & 10992              & 16                 &    class 4                                  &        (10.4\%)    \\
  shuttle     & 85849              & 9                  &      classes $\neq$ 1 (class 4 removed)     &  (7.17\%)          \\
  spambase    & 4601               & 57                 &    spam                                     &           (39.4\%) \\
  \bottomrule
\end{tabular}
}
\end{table*}



The characteristics of these reference datasets are summarized in Table~\ref{table:data}. They are all available on the UCI repository \cite{Lichman2013} and the preprocessing is done in a classical way. 
We removed all non-continuous attributes as well as attributes taking less than $10$ differents values.
The \emph{http} and \emph{smtp} datasets belong to the KDD Cup '99 dataset \cite{KDD99,Tavallaee2009}, which consists of a wide variety of hand-injected  attacks (anomalies) in a closed network (normal background). They are classicaly obtained as described in \cite{Yamanishi2000}. These datasets are available on the \emph{scikit-learn} library \cite{sklearn2011}.
The \emph{shuttle} dataset is the fusion of the training and testing datasets available in the UCI repository. As in \cite{Liu2008}, we use instances from all different classes but class $4$. 
In the \emph{forestcover} data, the normal data are the instances from class~$2$ while instances from class $4$ are anomalies (as in \cite{Liu2008}). 
The \emph{ionosphere} dataset differentiates `good' from `bad' radars, considered here as abnormal. A `good' radar shows evidence of some type of structure in the ionosphere. A `bad' radar does not, its signal passing through the ionosphere.
The \emph{spambase} dataset consists of spam or non-spam emails. The former constitute the abnomal class.
The \emph{annthyroid} medical dataset on hypothyroidism contains one normal class and two abnormal ones, which form the outlier set.
The \emph{arrhythmia} dataset reflects the presence and absence (class $1$) of cardiac arrhythmia. The number of attributes being large considering the sample size, we removed attributes containing missing data.
The \emph{pendigits} dataset contains 10 classes corresponding to the digits from 0 to 9, examples being handwriting samples. As in \cite{Schubert2012}, the abnormal data are chosen to be those from class 4.
The \emph{pima} dataset consists of medical data on diabetes. Patients suffering from diabetes (positive class) were considered outliers.
The \emph{wild} dataset involves detecting diseased trees in Quickbird imagery. Diseased trees (class `w') is the abnormal class.
In the \emph{adult} dataset, the goal is to predict whether income exceeds \$ 50K/year based on census data. Only the 6 continuous attributes are kept.

\subsection{complementary results}
\begin{figure*}[!ht]
  \centering
  \includegraphics[width=\linewidth]{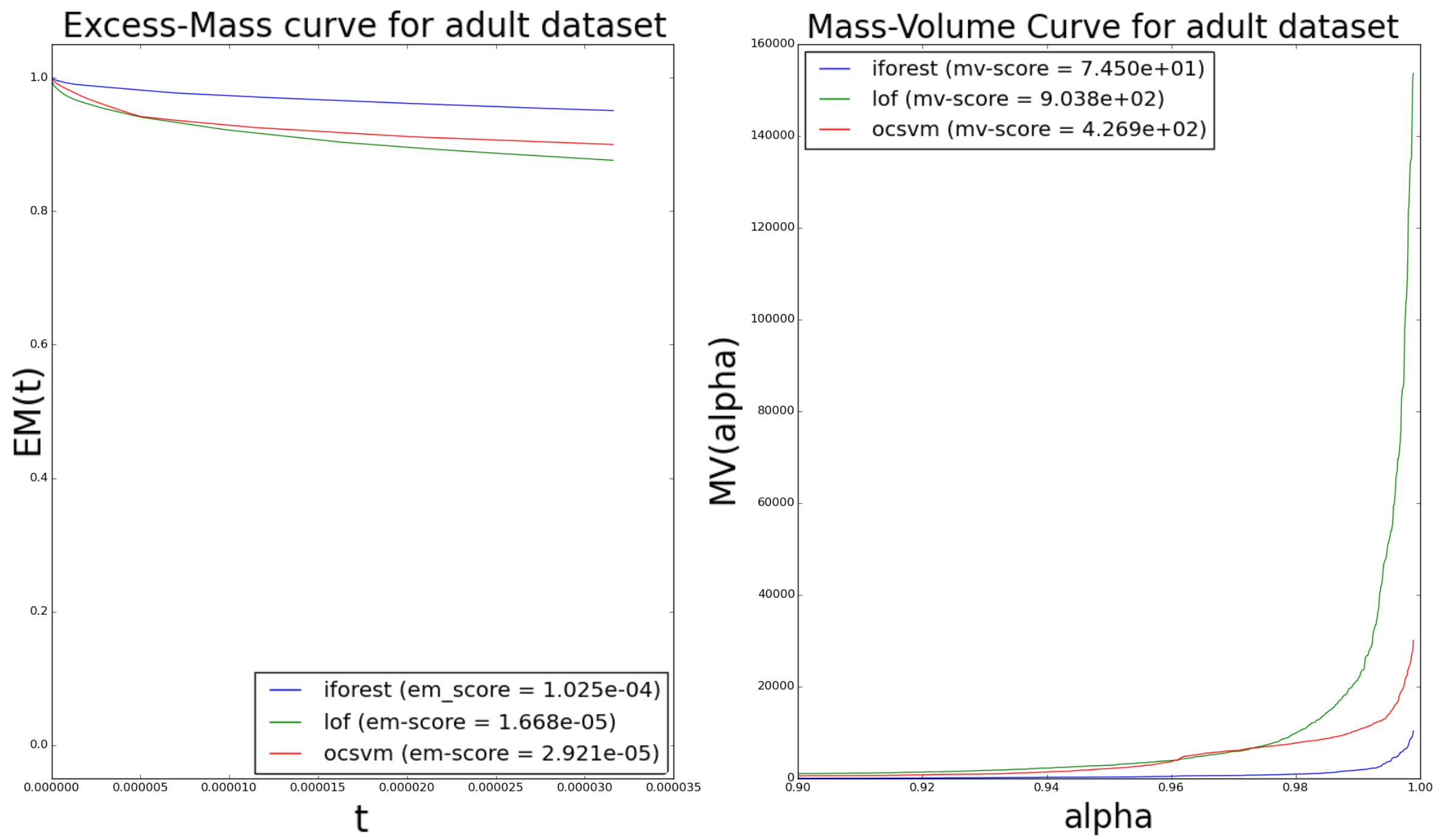}
\label{mv_em_adult}
  \caption{MV and EM curves for adult dataset (novelty detection framework). We can see that both in terms of EM and MV curves, iForest is found to perform better than OCSVM, which is itself found to perform better than LOF. Comparing to Table~\ref{table:results-semisupervised}, ROC and PR AUCs give the same ranking (iForest on adult $\succ$ OCSVM on adult $\succ$ LOF on adult). The 3 pairwise comparisons (iForest on adult, LOF on adult), (OCSVM on adult, LOF on adult) and (OCSVM on adult, iForest on adult) are then similarly ordered by EM, PR, MV and EM criteria.}
\end{figure*}

Results from the unsupervised framework (training and testing data are polluted by outliers) are similar for both EM and MV criteria. We just observe a slight decrease in accuracy.
Considering all the pairs, one observes $26/36 = 72\%$ (resp. $27/36 = 75\%$) of good comparisons \wrt~ROC-score (resp. \wrt~PR score) for EM, and $75\%$ (resp. $78\%$) of good comparisons \wrt~ROC-score (resp. \wrt~PR score) for MV.
Considering the pairs which are similarly ordered by ROC and PR criteria, the rate for EM as for MV increases to $24/31 = 77\%$.
\begin{table*}[!ht]
\centering
\caption{Results for the unsupervised setting still remains good: one can see that ROC, PR, EM, MV often do agree on which algorithm is the best (in bold), which algorithm is the worse (underlined) on some fixed datasets. When they do not agree, it is often because ROC and PR themselves do not, meaning that the ranking is not clear.}
\tabcolsep=0.11cm
\resizebox{\linewidth}{!} {
\begin{tabular}{l cccc c cccc c cccc}
\toprule
Dataset      & \multicolumn{4}{c}{iForest} & & \multicolumn{4}{c}{OCSVM} & & \multicolumn{4}{c}{LOF} \\ 
  \cmidrule{1-15}
~            & ROC  & PR   & EM    &  MV  &  & ROC   & PR    & EM    & MV    &    & ROC  & PR   & EM    & MV    \\
adult        &\bf 0.644 &\bf 0.234 &\bf 6.6e-05&\bf 2.7e02 & &0.627  &0.184  &1.8e-05&5.6e02   &  &\underline{0.545} &\underline{0.098} &\underline{7.4e-06}&\underline{1.9e03} \\
http         &\bf 0.999 &\bf 0.686 &1.4e-03&2.2e01&  &0.994  &0.207  &\bf 5.7e-03&\bf 3.3    &    &\underline{0.354} &\underline{0.019} &\underline{9.8e-05}&\underline{3.9e02} \\
pima         &\bf 0.747 &0.205 &\bf 1.2e-06&\bf 1.2e04 & &0.742  &\bf 0.211  &6.0e-07&1.9e04  &   &\underline{0.686} &\underline{0.143} &\underline{6.0e-07}&\underline{3.2e04} \\
smtp         &0.902 &\underline{0.004} &\underline{2.7e-04}&\underline{8.6e01}&  &\underline{0.852}  &\bf 0.365  &\bf 1.4e-03&7.7   &     &\bf 0.912 &0.057 &1.1e-03&\bf 7.0    \\ 
wilt         &0.443 &0.044 &3.7e-05&\underline{2.2e03} & &\underline{0.318}  &\underline{0.036}  &\bf 3.9e-05&\bf 4.3e02 &    &\bf 0.620 &\bf 0.066 &\underline{2.0e-05}&8.9e02 \\  &&&&&&&&&&&&&& \\

annthyroid   &\bf 0.820 &\bf 0.309 &\bf 6.9e-05&7.7e02 & &\underline{0.682}  &0.187  &4.1e-05&\bf 3.1e02  &   &0.724 &\underline{0.175} &\underline{1.6e-05}&\underline{4.1e03} \\
arrhythmia   &\bf 0.740 &0.416 &\bf 8.4e-05&\bf 1.1e02 & &0.729  &\bf 0.447  &6.8e-05&1.2e02   &  &\underline{0.729} &\underline{0.409} &\underline{5.6e-05}&\underline{1.5e02} \\
forestcov.   &0.882 &0.062 &\underline{3.2e-05}&\underline{2.3e02} & &\bf 0.951  &\bf 0.095  &4.4e-05&1.4e02    & &\underline{0.542} &\underline{0.016} &\bf 2.4e-04 &\bf 4.6e01 \\
ionosphere   &\underline{0.895} &\underline{0.543} &7.4e-05&\underline{9.3e01} & &\bf 0.977  &\bf 0.903  &\bf 8.7e-05&\bf 7.7e01   &  &0.969 &0.884 &\underline{6.9e-05}&1.0e02 \\
pendigits    &0.463 &0.077 &2.7e-04&2.5e01 & &\underline{0.366}  &\underline{0.067}  &\underline{2.6e-04}&\underline{2.8e01}&   &\bf 0.504 &\bf 0.089 &\bf 4.5e-04&\bf 1.6e01 \\
shuttle      &\bf 0.997 &\bf 0.979 &7.1e-07&1.2e05 & &0.992  &0.904  &\bf 5.8e-06&\bf 1.7e02  &   &\underline{0.526} &\underline{0.116} &\underline{7.1e-07}&\underline{1.7e07} \\
spambase     &\bf 0.799 &\bf 0.303 &\bf 2.2e-04&\bf 3.5e01 & &0.714  &0.214  &1.5e-04&2.9e02  &   &\underline{0.670} &\underline{0.129} &\underline{3.7e-05}&\underline{2.7e04} \\
\bottomrule
\end{tabular}
}
\end{table*}
Figure~\ref{mv_em_adult} shows excess-mass and mass-volume curves on the adult dataset in a novelty detection setting. 
Corresponding figures for the other datasets follow.

\clearpage

\begin{figure*}[!ht]
\label{mv_em_http}
  \centering
  \caption{MV and EM curves for http dataset (novelty detection framework)}
  \includegraphics[trim=172 52 165 70, clip, width=\linewidth]{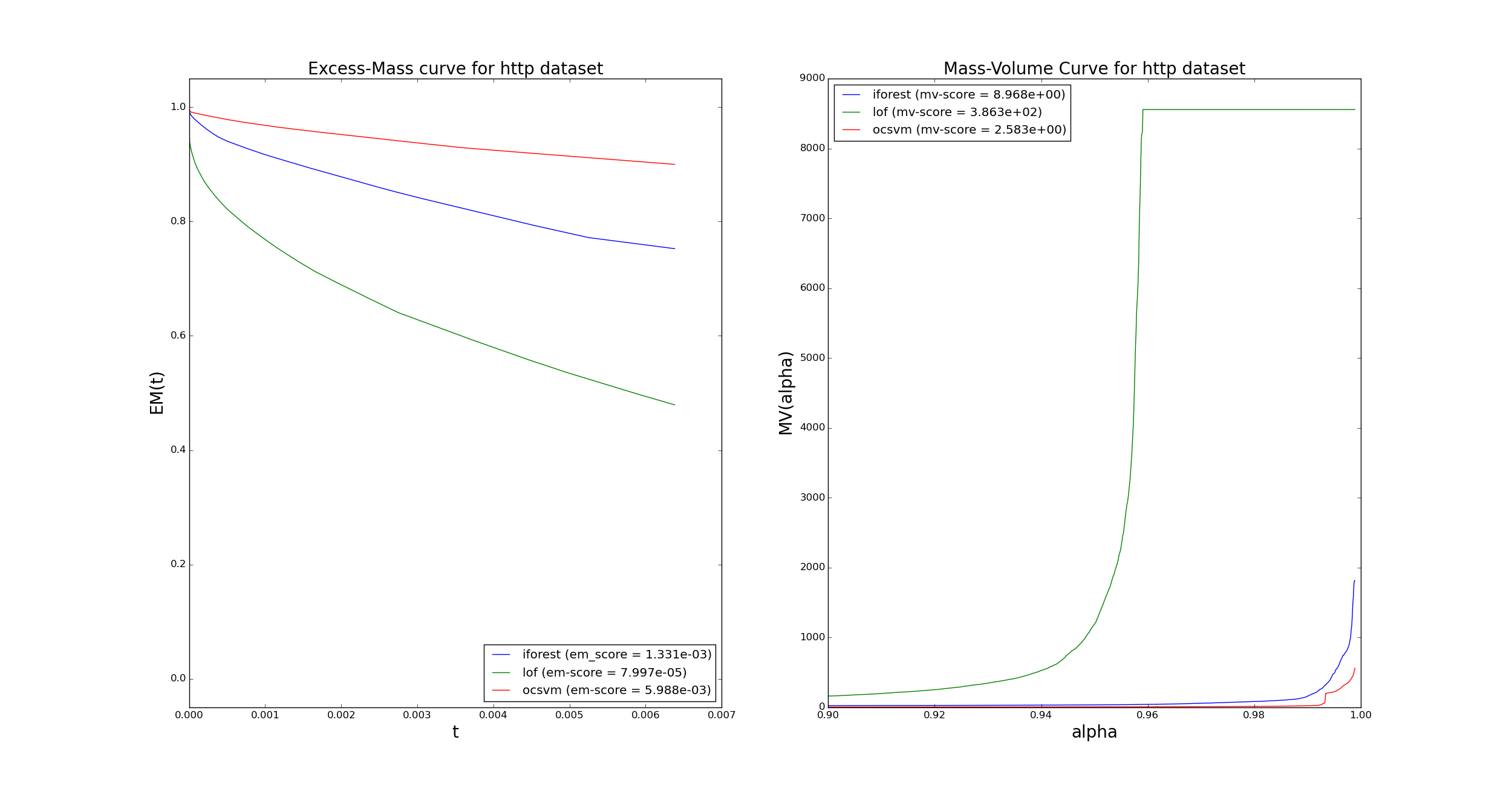}
\end{figure*}
\begin{figure*}[!ht]
\label{mv_em_http_unsupervised}
  \centering
  \caption{MV and EM curves for http dataset (unsupervised framework)}
  \includegraphics[trim=172 52 165 70, clip, width=\linewidth]{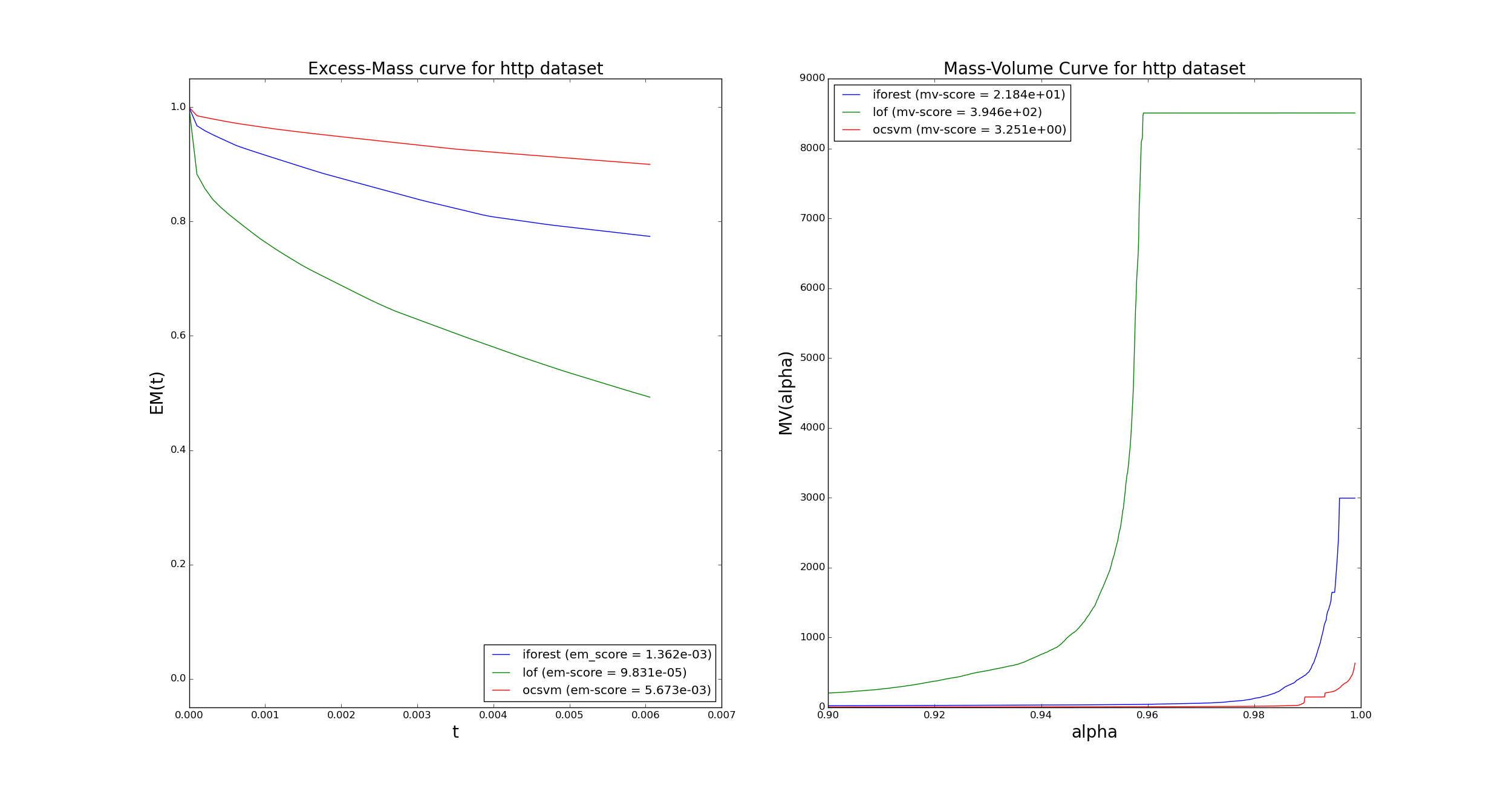}
\end{figure*}


\begin{figure*}[!ht]
\label{mv_em_pima}
  \centering
  \caption{MV and EM curves for pima dataset (novelty detection framework)}
  \includegraphics[trim=172 52 165 70, clip, width=\linewidth]{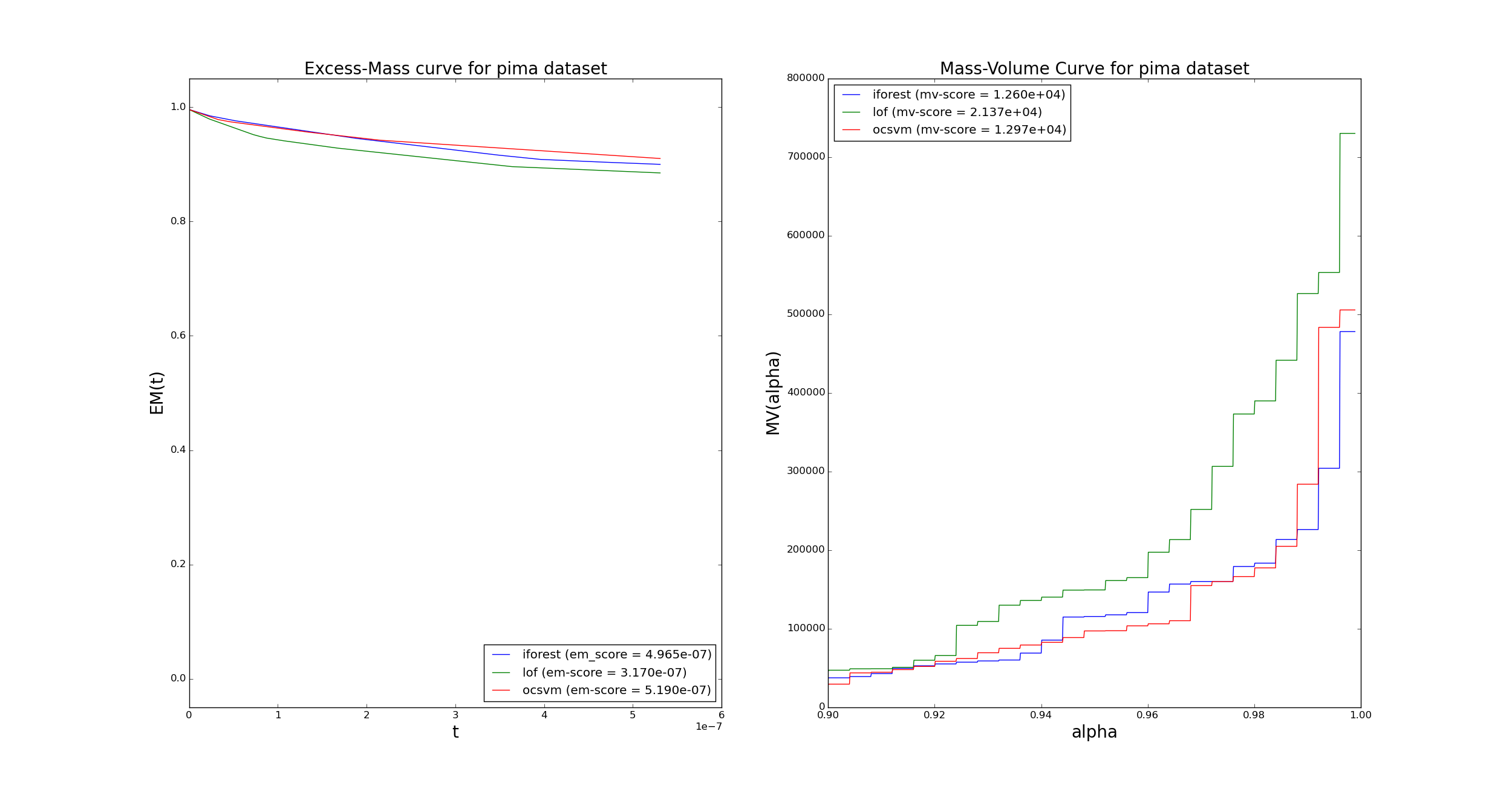}
\end{figure*}
\begin{figure*}[!ht]
\label{mv_em_pima_unsupervised}
  \centering
  \caption{MV and EM curves for pima dataset (unsupervised framework)}
  \includegraphics[trim=172 52 165 70, clip, width=\linewidth]{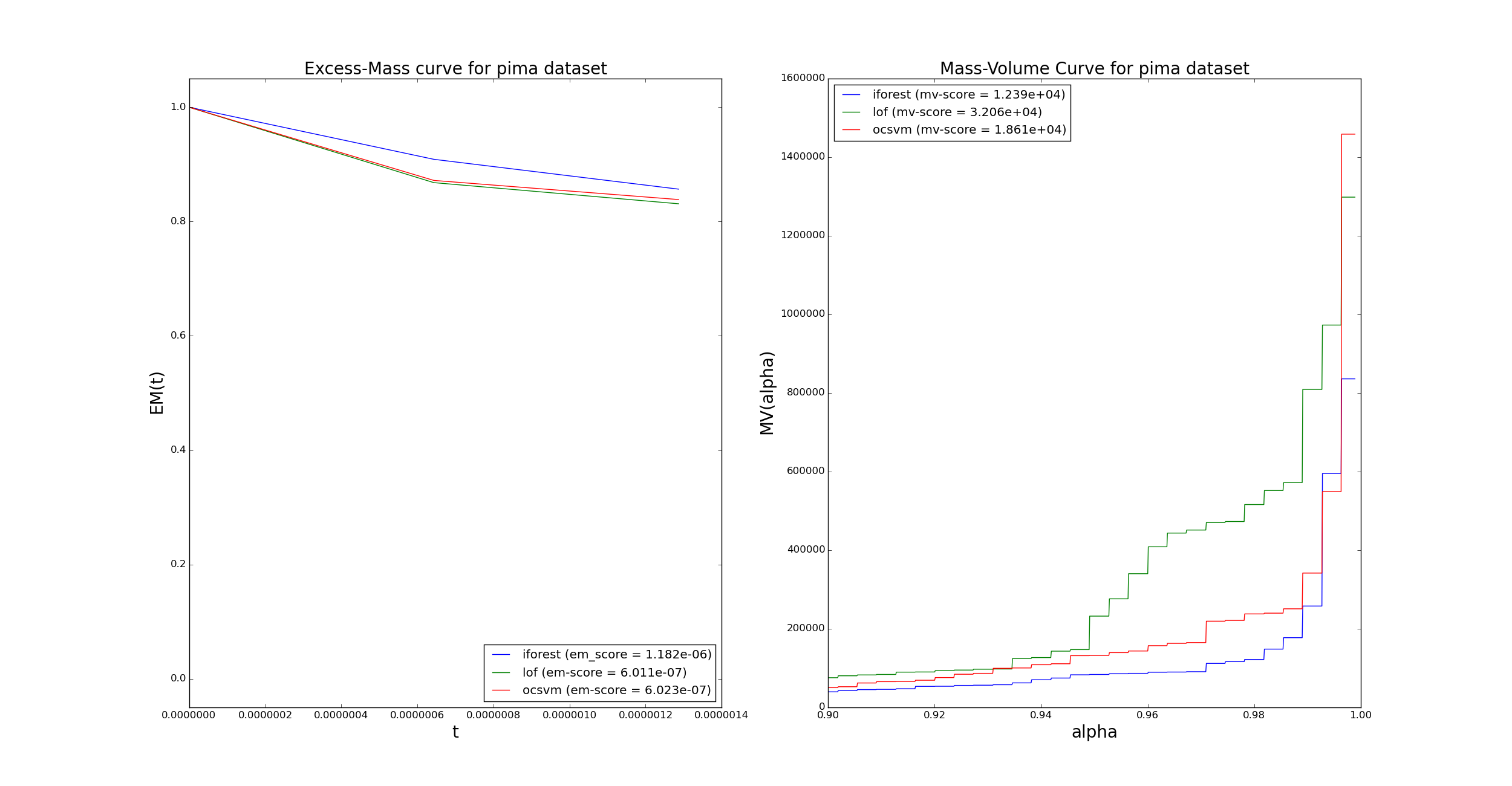}
\end{figure*}

\begin{figure*}[!ht]
\label{mv_em_smtp}
  \centering
  \caption{MV and EM curves for smtp dataset (novelty detection framework)}
  \includegraphics[trim=172 52 165 70, clip, width=\linewidth]{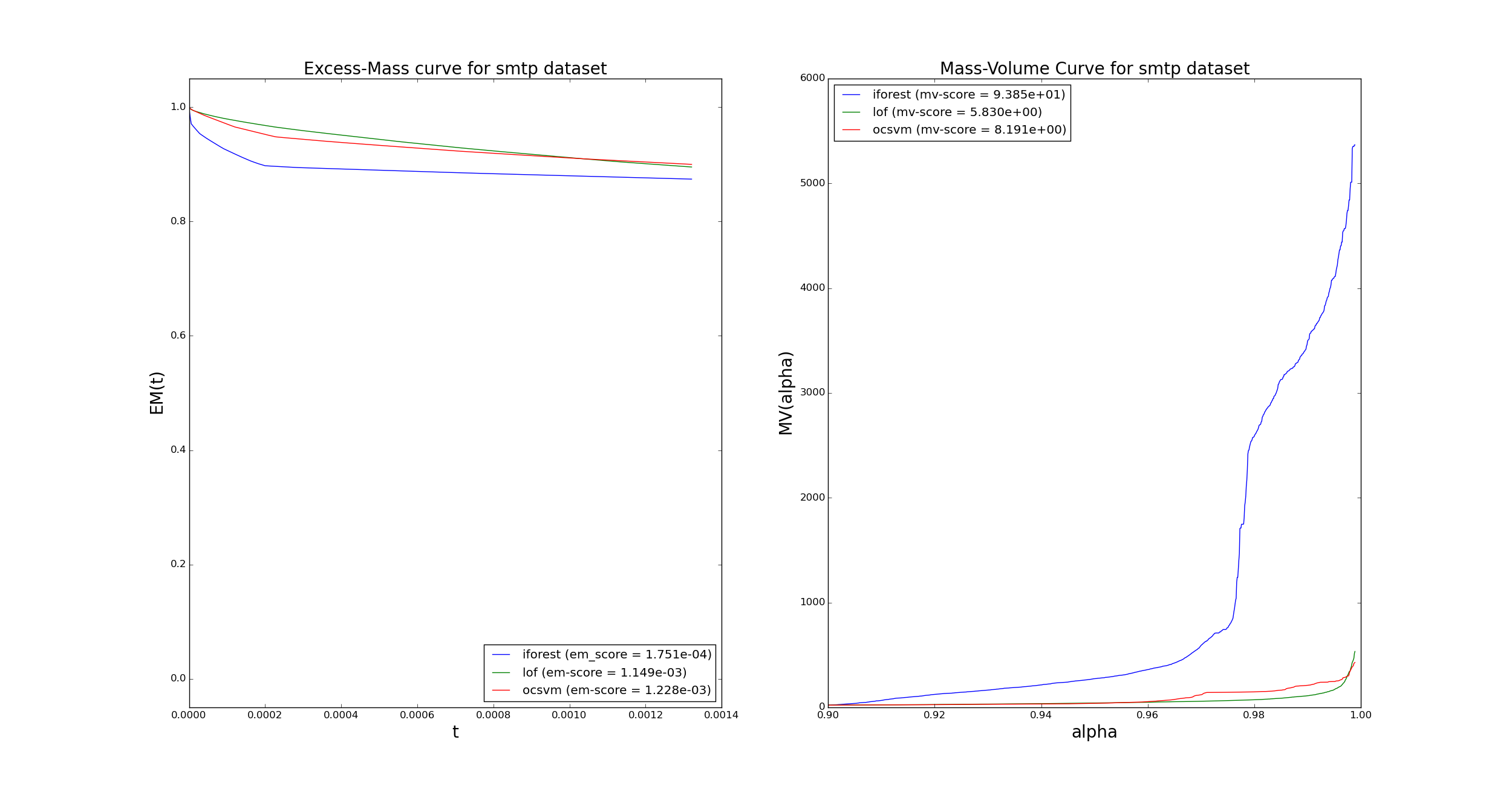}
\end{figure*}
\begin{figure*}[!ht]
\label{mv_em_smtp_unsupervised}
  \centering
  \caption{MV and EM curves for smtp dataset (unsupervised framework)}
  \includegraphics[trim=172 52 165 70, clip, width=\linewidth]{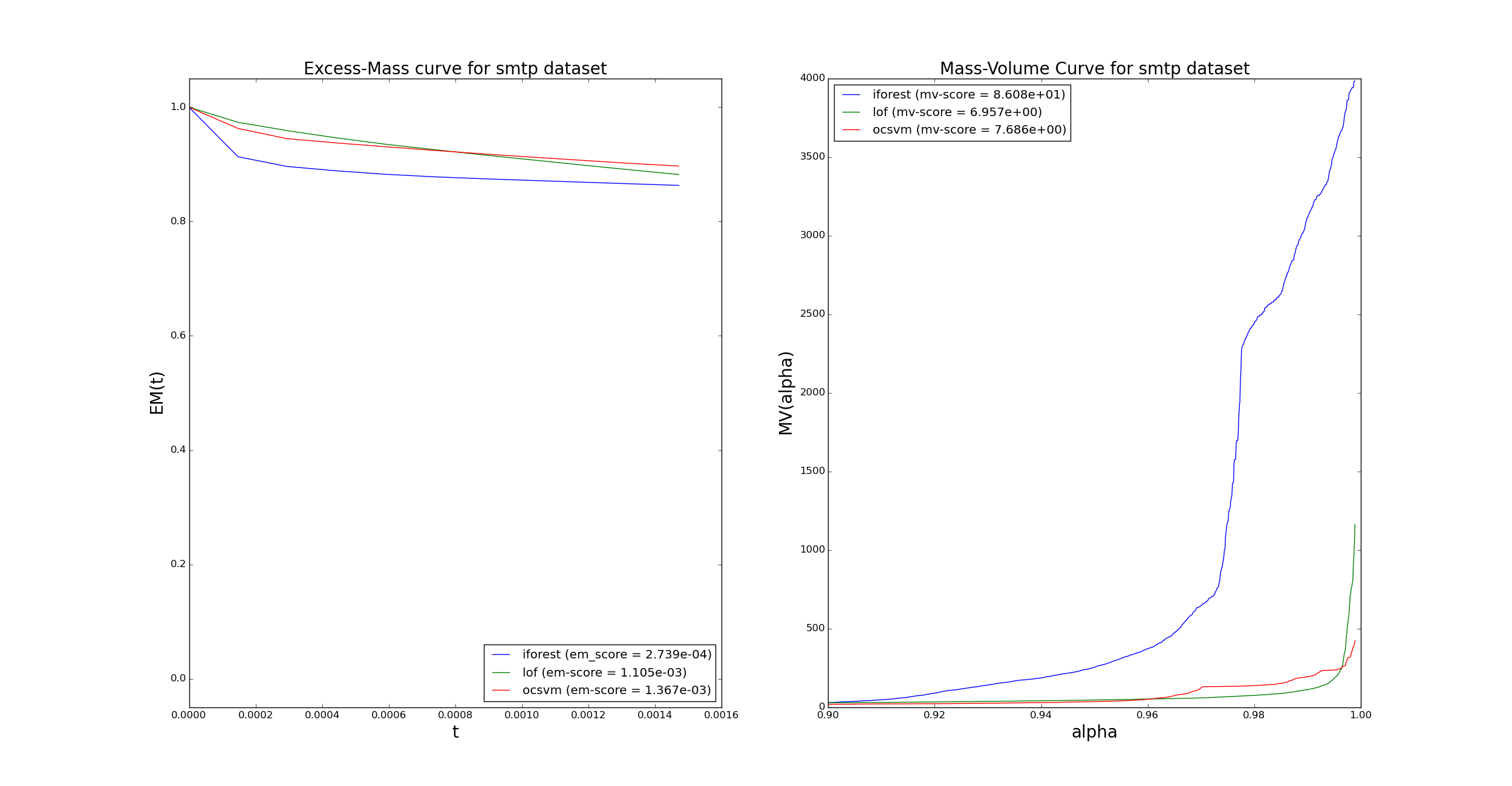}
\end{figure*}

\begin{figure*}[!ht]
\label{mv_em_wilt}
  \centering
  \caption{MV and EM curves for wilt dataset (novelty detection framework)}
  \includegraphics[trim=172 52 165 70, clip, width=\linewidth]{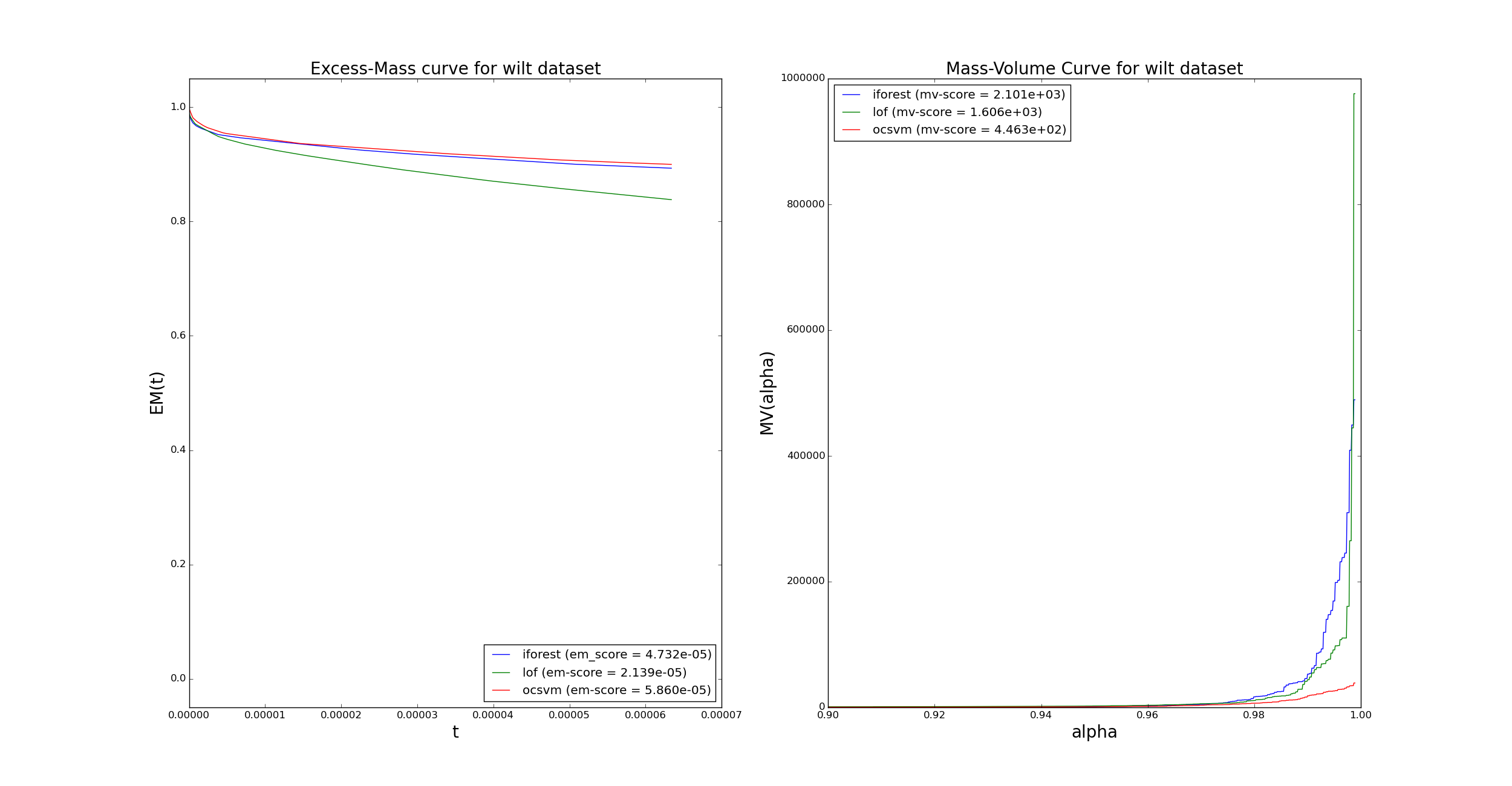}
\end{figure*}
\begin{figure*}[!ht]
\label{mv_em_wilt_unsupervised}
  \centering
  \caption{MV and EM curves for wilt dataset (unsupervised framework)}
  \includegraphics[trim=172 52 165 70, clip, width=\linewidth]{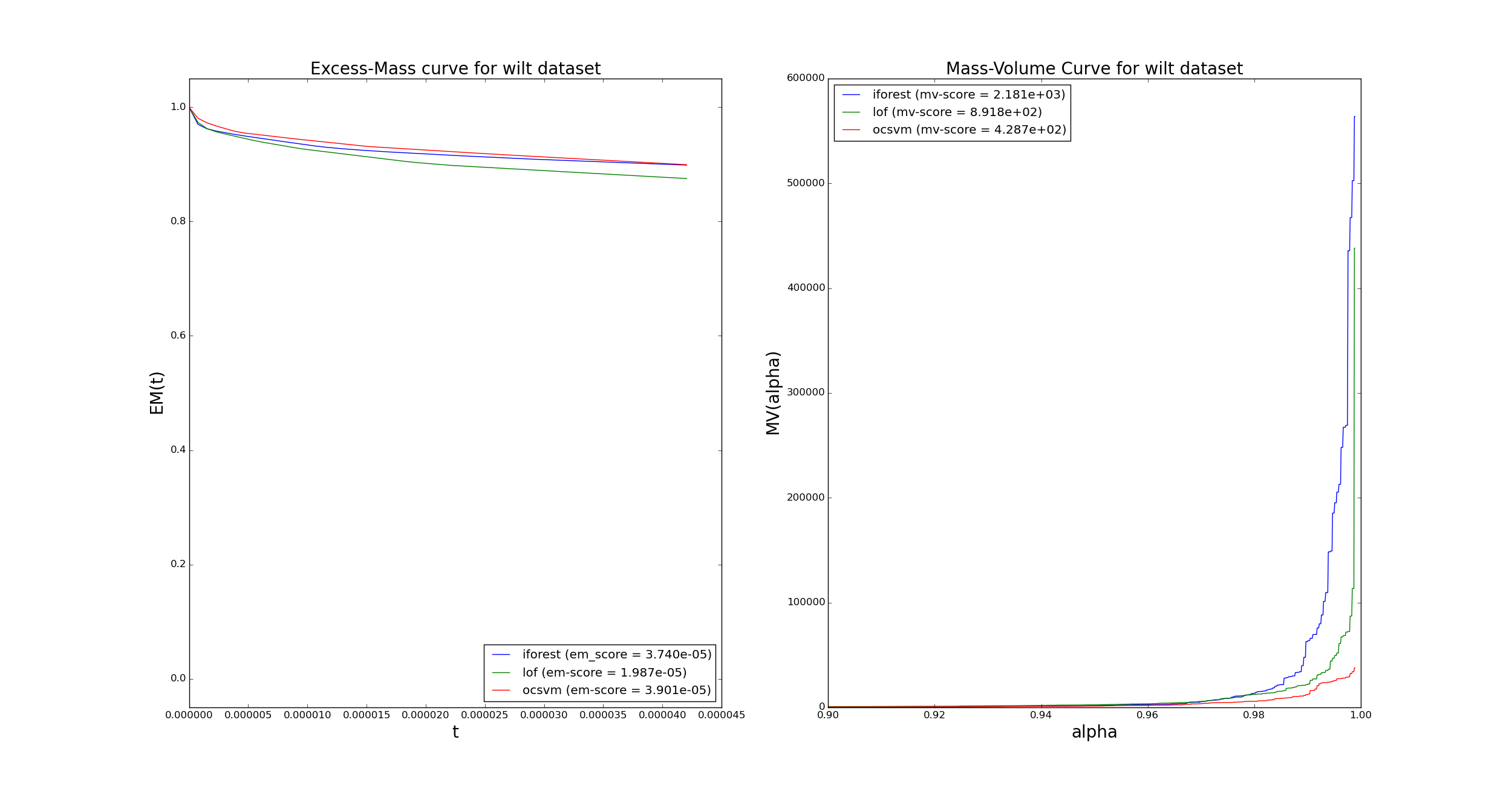}
\end{figure*}

\begin{figure*}[!ht]
  \centering
  \caption{MV and EM curves for adult dataset (novelty detection framework).}
  \includegraphics[trim=172 52 165 70, clip, width=\linewidth]{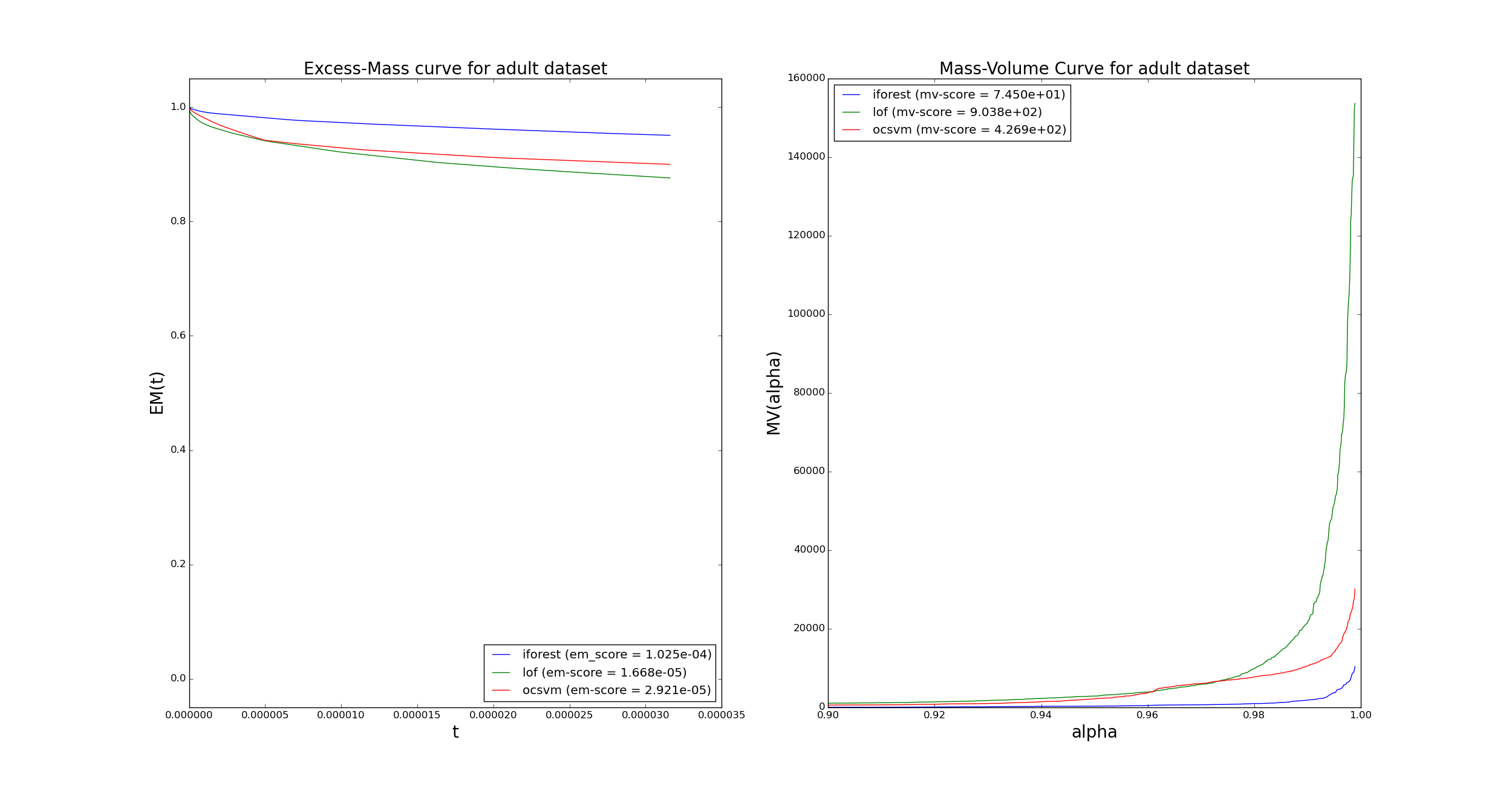}
\end{figure*}

\begin{figure*}[!ht]
  \centering
  \caption{MV and EM curves for adult dataset (unsupervised framework)}
  \includegraphics[trim=172 52 165 70, clip, width=\linewidth]{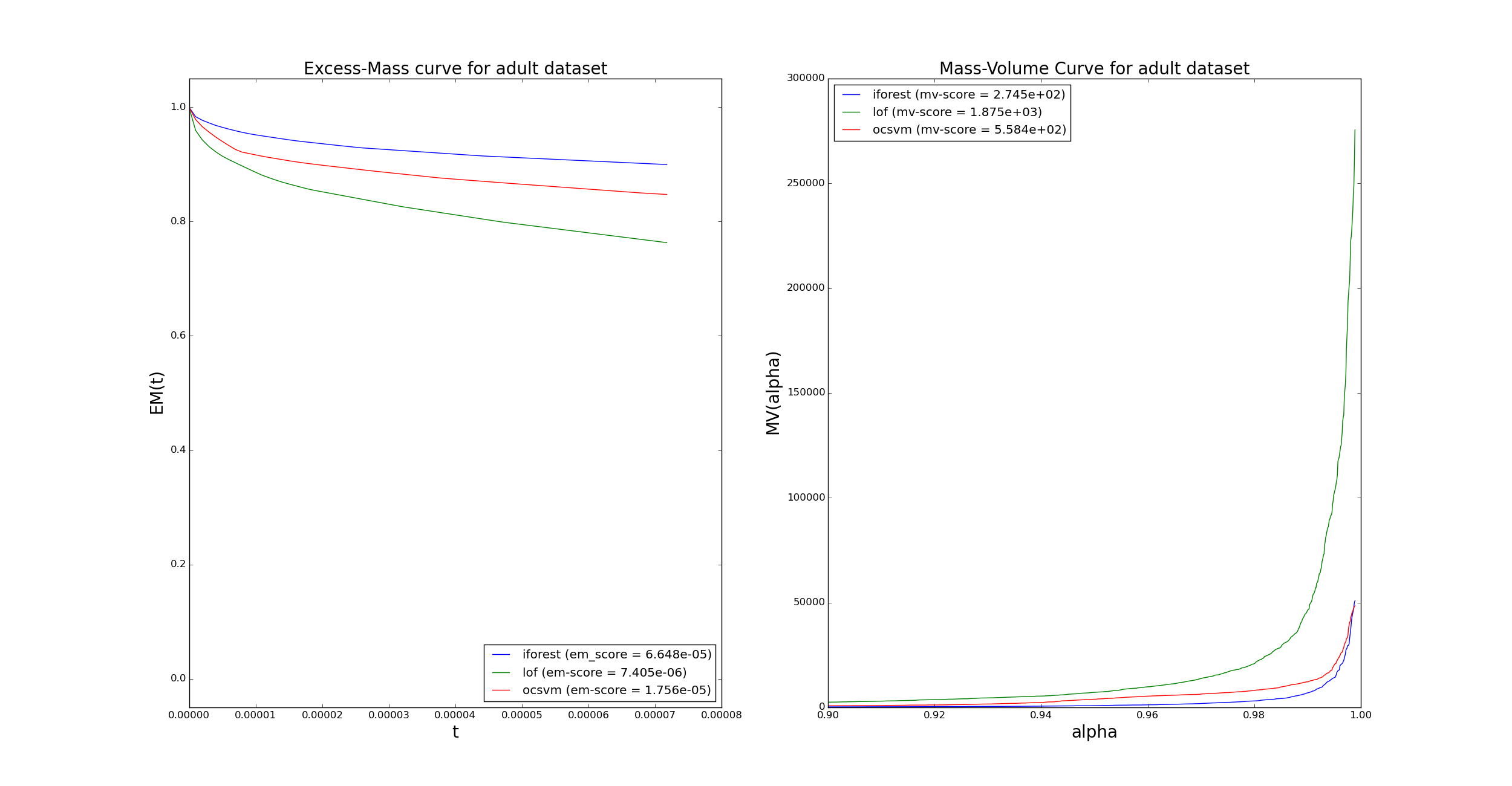}
\label{mv_em_adult_unsupervised}
\end{figure*}

\end{document}